\documentclass[10pt,twocolumn,letterpaper]{article}

\usepackage[pagenumbers]{cvpr}              

\usepackage{times}
\usepackage{graphicx}
\usepackage{amsmath}
\usepackage{amssymb}
\usepackage{booktabs}
\usepackage{comment}
\usepackage{cases}
\usepackage{setspace}
\usepackage{overpic}
\usepackage{rotating}
\usepackage{subeqnarray}
\usepackage{multirow}
\usepackage{tabularx}
\usepackage{epstopdf}
\usepackage{tikz}
\usepackage{enumerate}
\usepackage{here}
\usepackage{bm}
\usepackage{color}
\usepackage{xcolor}
\usepackage{colortbl,booktabs}
\usepackage{bbm}
\usepackage[ruled,linesnumbered]{algorithm2e}
\usepackage{setspace}
\usepackage{tabularx}
\usepackage{diagbox}
\usepackage{url}
\usepackage[accsupp]{axessibility}  

\usepackage[pagebackref,breaklinks,colorlinks]{hyperref}

\usepackage[capitalize]{cleveref}
\crefname{section}{Sec.}{Secs.}
\Crefname{section}{Section}{Sections}
\Crefname{table}{Table}{Tables}
\crefname{table}{Tab.}{Tabs.}

\def\cvprPaperID{27} 
\def\confName{CVPR}
\def\confYear{2023}

\begin{document}

\title{Rip Current Segmentation: A Novel Benchmark and YOLOv8 Baseline Results}

\author{Andrei Dumitriu$^{1, 2}$, Florin Tatui$^{2}$, Florin Miron$^{2}$,  Radu Tudor Ionescu$^{2}$, Radu Timofte$^{1}$ \\
$^{1}$Computer Vision Lab, CAIDAS \& IFI, University of Würzburg, Germany\\
$^{2}$University of Bucharest, Romania\\
{\tt\small {andrei.dumitriu}@uni-wuerzburg.de}\\
}

\twocolumn[{%
\renewcommand\twocolumn[1][]{#1}%
\maketitle
\begin{center}
    \centering
    \captionsetup{type=figure}
    \includegraphics[width=0.2\textwidth]{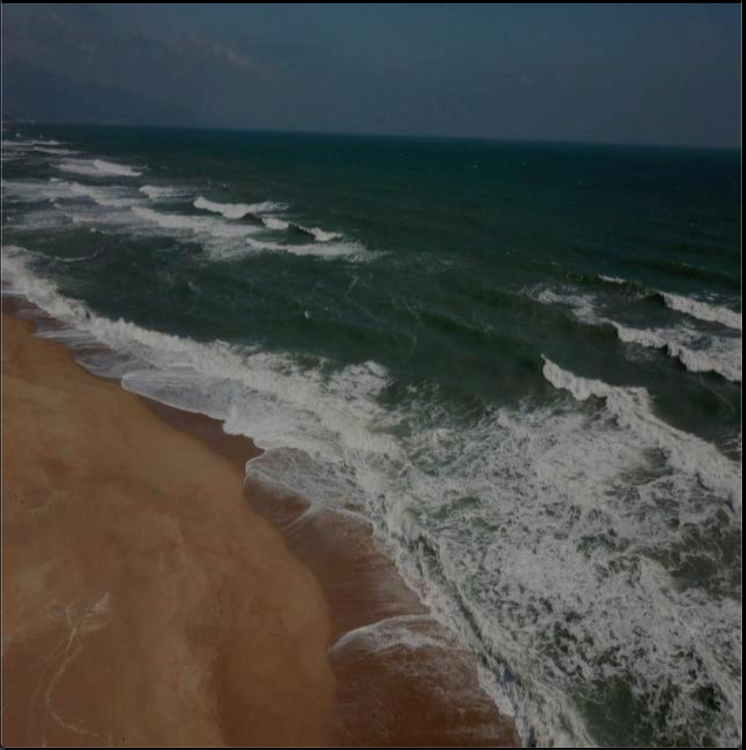}
    \includegraphics[width=0.2\textwidth]{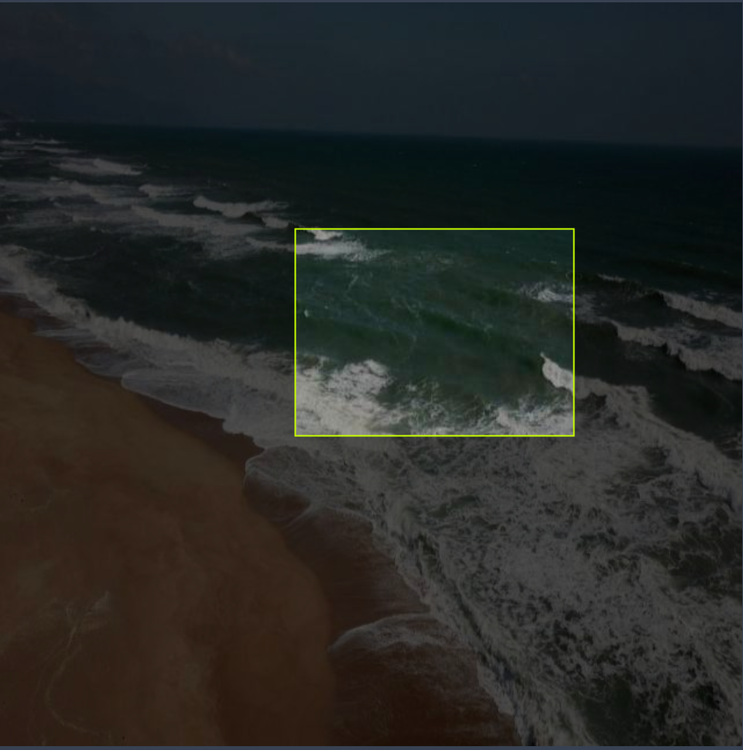}
    \includegraphics[width=0.2\textwidth]{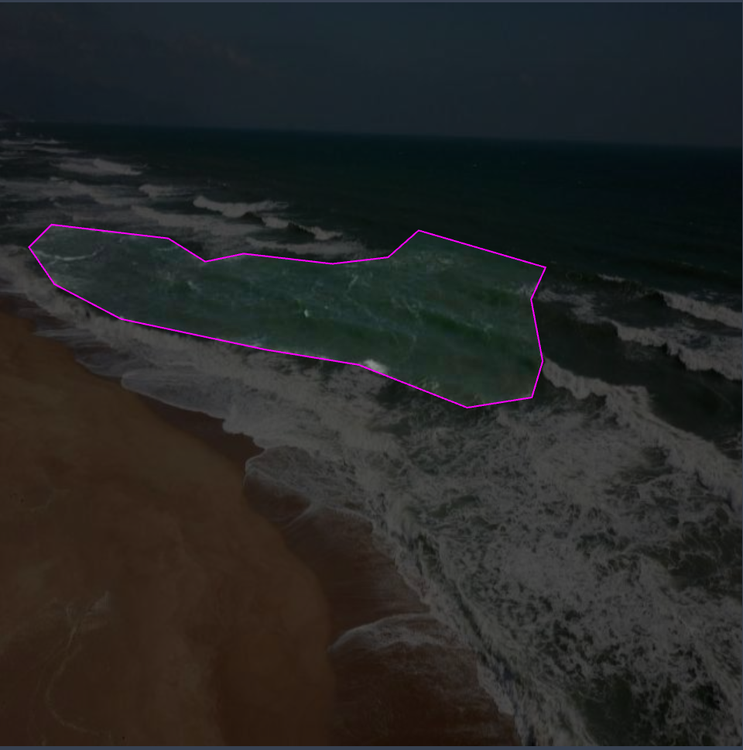}
    \includegraphics[width=0.2\textwidth]{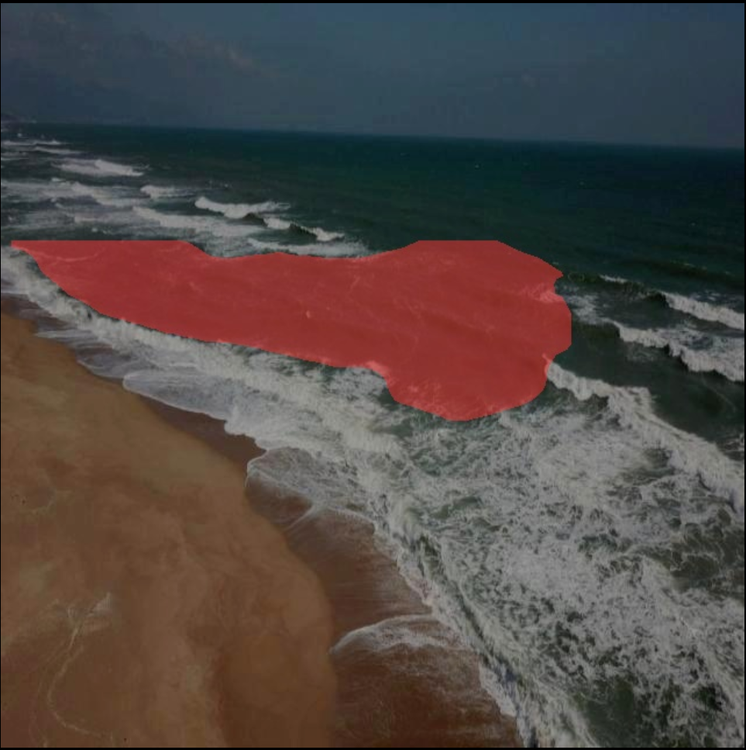}
    \captionof{figure}{An illustration of a visible rip current and its detection task, displayed from left to right:
    1. The original photo,
    2. The bounding box annotation from the YOLO-Rip dataset,
    3. Our ground-truth annotation for instance segmentation,
    4. Prediction using YOLOv8. This example highlights how bounding boxes may exclude relevant parts of the rip currents while also incorporating surrounding noise.}
\end{center}%
}]

\begin{abstract}
\vspace{-0.2cm}
Rip currents are the leading cause of fatal accidents and injuries on many beaches worldwide, emphasizing the importance of automatically detecting these hazardous surface water currents. In this paper, we address a novel task: rip current instance segmentation. We introduce a comprehensive dataset containing 2,466 images with newly created polygonal annotations for instance segmentation, used for training and validation. Additionally, we present a novel dataset comprising 17 drone videos (comprising about 24K frames) captured at 30 FPS, annotated with both polygons for instance segmentation and bounding boxes for object detection, employed for testing purposes. We train various versions of YOLOv8 for instance segmentation on static images and assess their performance on the test dataset (videos). The best results were achieved by the YOLOv8-nano model (runnable on a portable device), with an mAP50 of 88.94\% on the validation dataset and 81.21\% macro average on the test dataset. The results provide a baseline for future research in rip current segmentation. Our work contributes to the existing literature by introducing a detailed, annotated dataset, and training a deep learning model for instance segmentation of rip currents. The code, training details and the annotated dataset are made publicly available at \url{https://github.com/Irikos/rip_currents}.
\end{abstract}

\section{Introduction}
\label{sec:intro}
Rip currents are a common life-threatening hazard\cite{da2003analysis, lushine1991study, brewster2019estimations, brander2013brief} found at beaches bounded by large bodies of water, such as oceans, seas or large lakes. They can vary significantly in size and speed, being driven by changes in their controlling factors such as nearshore hydrodynamics and underwater morphology. While many are naturally occurring, some are the result of human activity, being related to the presence of coastal structures or other activities. The rip current's main danger is its outward pulling power, some of them reaching up to 8.7km/h, which is faster than an Olympic swimmer \cite{noaa2023ripcurrents}. Alongside the rip current's drag, the actual danger is increased by the lack of knowledge on most people's part, both in recognizing and properly reacting when caught in one. When trapped in a rip current, a common reaction is to panic and swim against it, resulting in fatigue and possible death. An advisable solution would be to swim perpendicular to it (parallel to the beach) and escape its grasp. An even better solution would be early identification of the rip current and not getting caught in it at all. 

Rip currents are studied worldwide using traditional methods such as ecological dye experiments, drifter measurements and current meter surveys or video/drone footage with human analysis \cite{castelle2016rip, benassai2017rip, leatherman2017techniques}. There have also been attempts of automatic rip current detection, using both traditional methods and deep learning.

The most relevant available dataset at the moment was started by de Silva~\etal\cite{desilva2021frcnn}, with 1,740 images with rips and 700 images without rips, collected mostly from Google Earth imagery. Alongside these, there are several videos with and without rip currents, for which bounding box annotations have not been provided. Later, Zhu~\etal\cite{zhu2022yolo} created a second dataset based on the images collected by de Silva, by discarding several of them and adding 1,352 high-resolution images (746 with rips and 606 without) from beaches along the coast of South China. The images are rescaled to a resolution of $640\times 640$ pixels and are annotated with axis-aligned bounding boxes to label the rip currents.

These datasets - and the methods applied on them - only produce bounding box detections at best. While these are good, the amorphous property of rip currents and the limited format of bounding box detection imply many situations where either information is left out or background information is added to the detection. Creating a segmented dataset, while with its own challenges, opens the possibility of precisely detecting a rip current's position, shape and, possibly, direction. Furthermore, rip current detection and segmentation can increase the awareness, education and overall reaction of swimmers, surfers and other beach goers.

To this end, we propose a novel rip current segmentation benchmark, that contains both images and videos with polygonal annotations  of rip currents. Our contribution can be summarized as:
\begin{itemize}
  \item Annotating 2,466 images with polygonal annotations.  
  \item Collecting 17 videos of rip currents (24,295 total frames) and annotating them with both bounding boxes and polygonal annotations.
  \item Training and comparing several versions of YOLOv8\cite{Jocher_YOLO_by_Ultralytics_2023} for rip current instance segmentation, serving as baselines for future research.
\end{itemize}

\begin{figure}[t!]
    \centering
    \includegraphics[width=\columnwidth,scale=0.5]{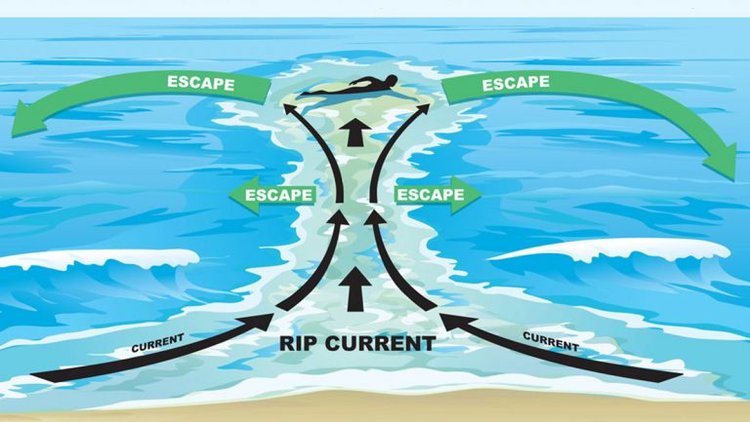}
    \caption[Rip current illustration]{Rip current illustration showing the direction of the current and how to swim in order to escape it. Source: \url{https://www.noaa.gov/}.}
    \label{fig:rip_current_intro}
\end{figure}

\section{Related Works}
\subsection{Rip Current Detection Methods}

Rip currents are widely studied in natural sciences\cite{leatherman2017techniques, sonu1972field, inman1980field, brander2000morphodynamics, dusek2013rip, zhang2021rip, valipour2018analytical}. Traditional observation methods include naked eye observations and camera monitoring systems\cite{prodger2012argus, holman2007history, dusek2019webcat}. While in-situ observations using floating objects or GPS drifters offer precision\cite{castelle2014rip, castelle2016rip, short1994rip}, these methods are costly, location-dependent, and impractical for detecting flash rips. Alternative methods like laser rangefinders and drones with tracer dye provide more flexibility and better views\cite{clark2014aerial, kim2021analysis, pujianiki2020application}. Our machine learning-based method surpasses these traditional approaches by being cost-effective, easily scalable, and capable of real-time detection, making it more accessible and practical for the general public.

Numerical models like SWAN\cite{booij1996swan, dudkowska2020rip}, FUNWAVE\cite{wang2018numerical, hong2021numerical}, SWASH\cite{wang2018numerical, chang2021mechanism}, and XBEACH\cite{roelvink2010xbeach, mouragues2021headland, dudkowska2020rip} can simulate rip currents based on environmental conditions\cite{edition2015rip, dusek2013probabilistic, arun2014rip}. These models help visualize rip current characteristics, but are not always precise representations due to the varied nature of rip currents. Our method offers real-time detection and could potentially be integrated with forecasting methods for improved accuracy and timely attention to high-risk periods.

Various methods analyze rip currents using collected video and image data \cite{prodger2012argus, holman2007history, dusek2019webcat, holland1997practical}, including time exposure images or timex images. These images, averaging video frames over several minutes, enhance visual rip current detection \cite{nelko2011rip, lippmann1989quantification}. However, they are limited to wave breaking rip currents and shifting positions. Maryan~\etal \cite{maryan2019machine} compared detection algorithms on timex images, but their dataset was too small for deep learning. Pitman~\etal  \cite{pitman2016synthetic} used synthetic rip current images, improving accuracy but underpredicting rip current channels. Liu~\etal \cite{liu2019lifeguarding} employed a threshold and HSV-based segmentation method with 83\% accuracy, but it is limited to rip currents with visible sediments. Our method uses a novel dataset and deep learning for more robust instance segmentation-based detection.

Optical flow has been utilized for rip current detection in various studies. Philip~\etal \cite{philip2016flow} applied the Lukas-Kanade optical flow algorithm to determine water flow direction and isolate rip currents. Despite increasing resilience to small camera movements, their method requires a stable platform and only detects the predominant rip current direction. Mori~\etal \cite{mori2022flow} modified flow visualization fields, improving rip current detection and clarity. However, it also demands a stable camera platform. McGill~\etal \cite{mcgill2022flow} used dense optical flows with the Farneb\"{a}ck method and timex images, achieving 67.3\% and 96.2\% accuracy in detecting rip currents and channels, respectively. Their method is slow, taking up to 30 minutes to process a 10-minute video, and depends on camera placement and beach morphology. Optical flow methods offer certain advantages, such as rip current detection in the absence of wave-breaking patterns and providing flow information for comparison with in-situ observations. However, they are best suited for specific fixed camera placements and orientations. Our method is more noise-resistant and adaptable to various camera placements.

\begin{figure*}
    \centering
    \setlength{\tabcolsep}{1pt}
    \begin{tabular}{c c c c c c}
         (a) & (b) & (c) & (d) & (e) \tabularnewline
         \begin{turn}{90} {\raggedright With rips (easy)} \end{turn}
         \includegraphics[width=0.19\textwidth]{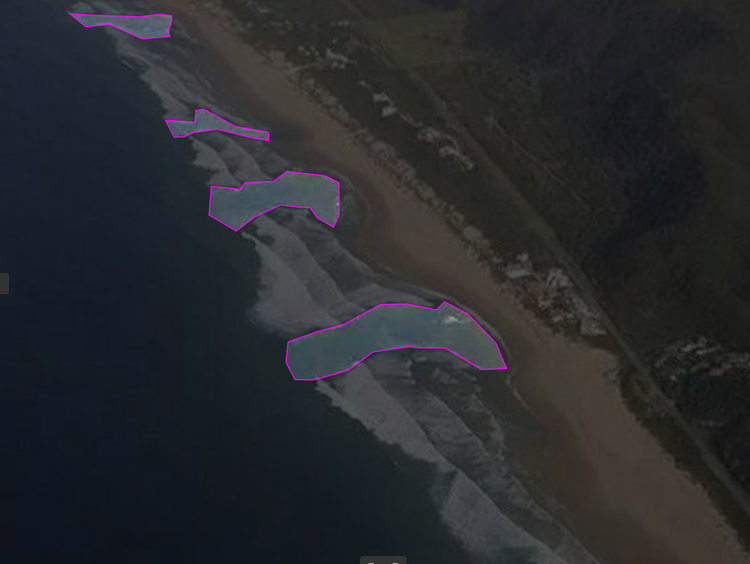} & 
         \includegraphics[width=0.19\textwidth]{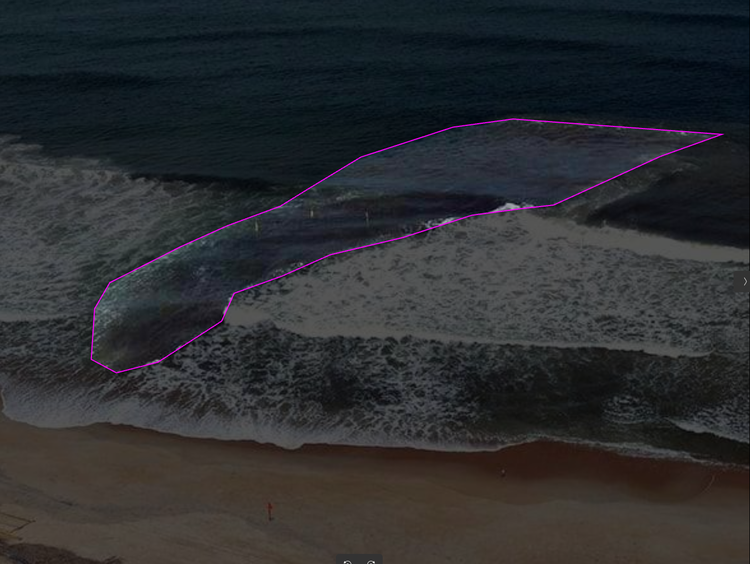} & 
         \includegraphics[width=0.19\textwidth]{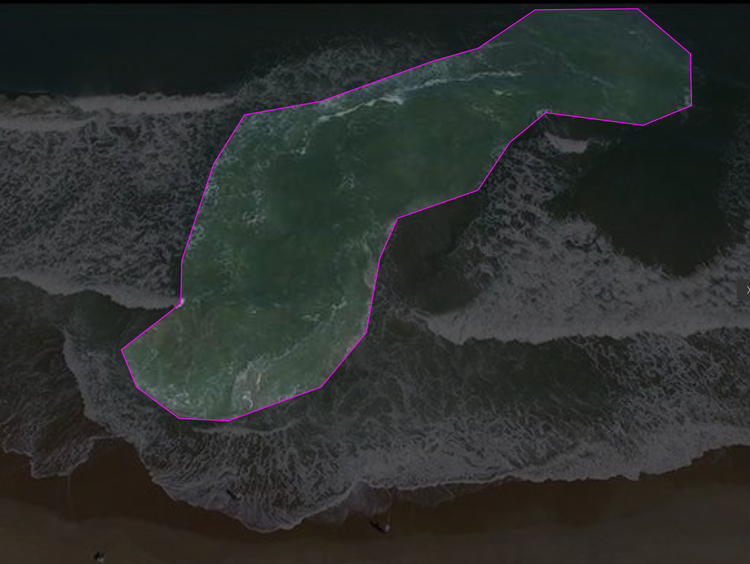} & 
         \includegraphics[width=0.19\textwidth]{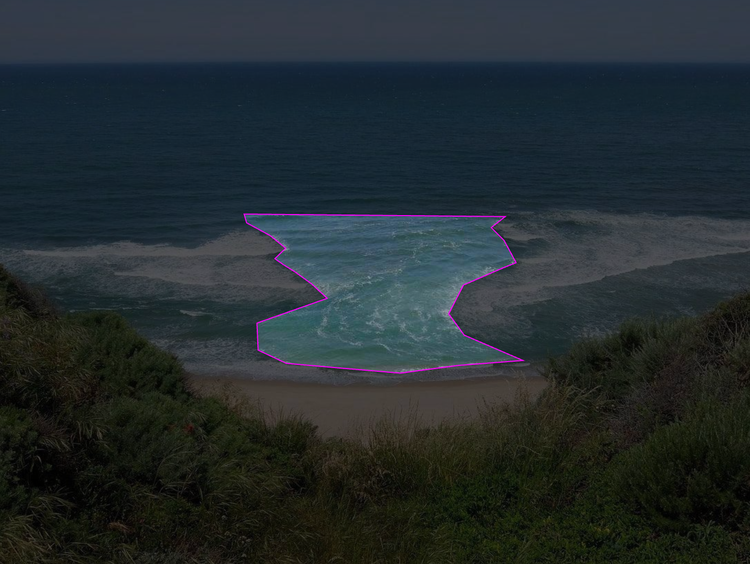} & 
         \includegraphics[width=0.19\textwidth]{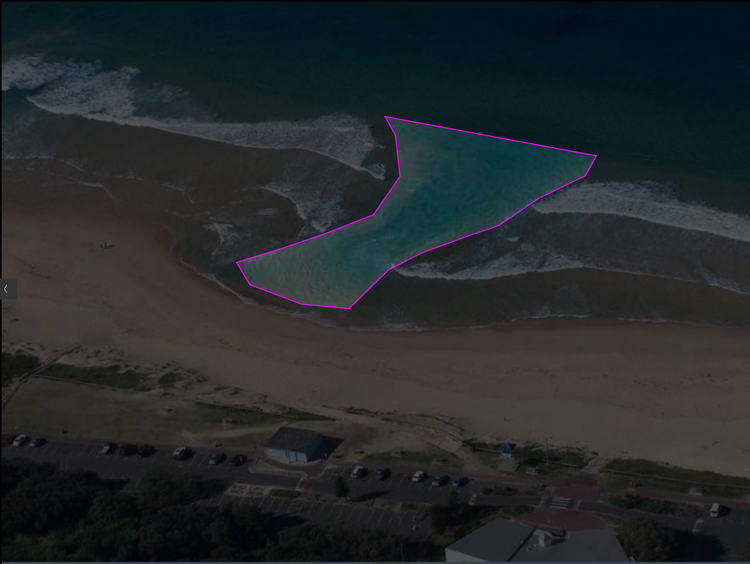} &  \tabularnewline
         \begin{turn}{90} {\raggedright With rips (hard)} \end{turn}
         \includegraphics[width=0.19\textwidth]{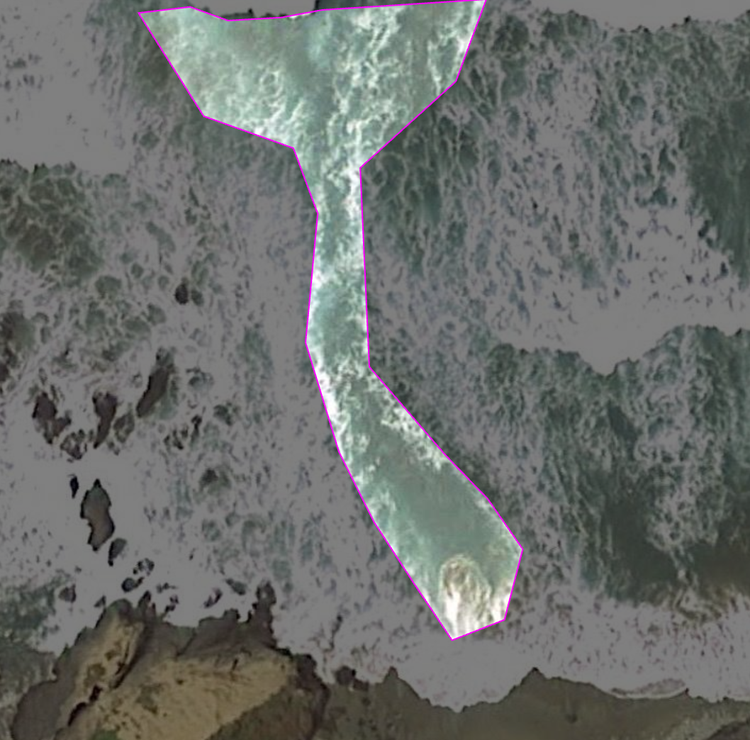} & 
         \includegraphics[width=0.19\textwidth]{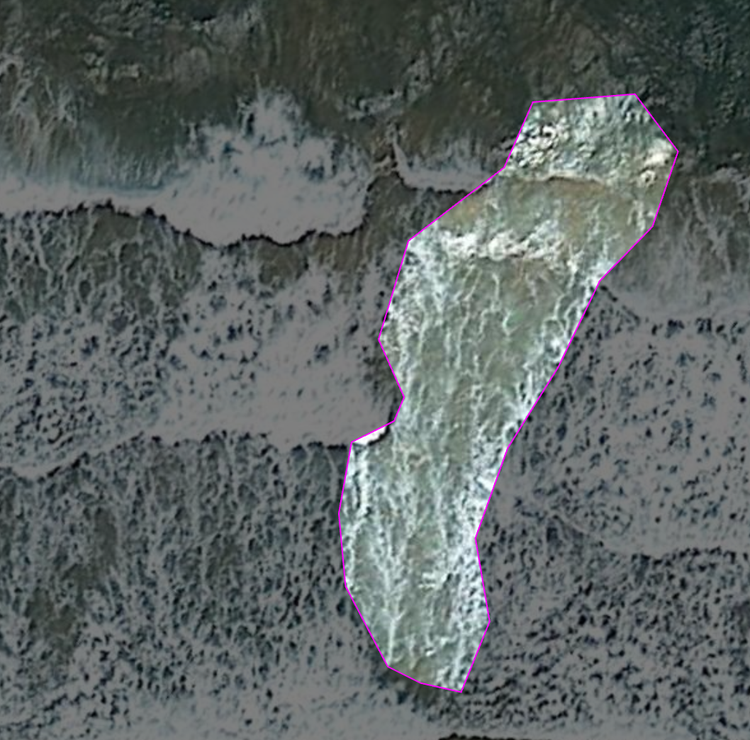} & 
         \includegraphics[width=0.19\textwidth]{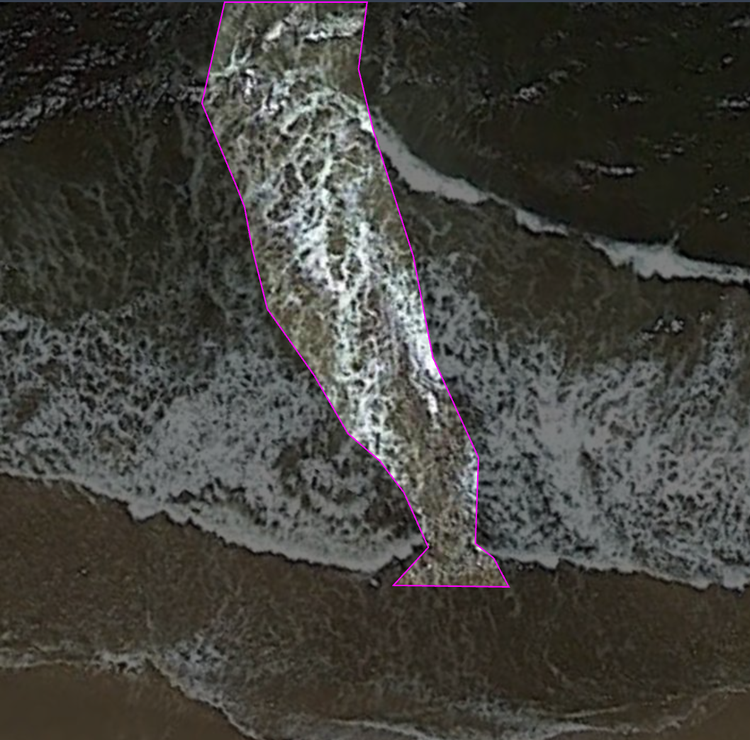} & 
         \includegraphics[width=0.19\textwidth, height=3.3cm]{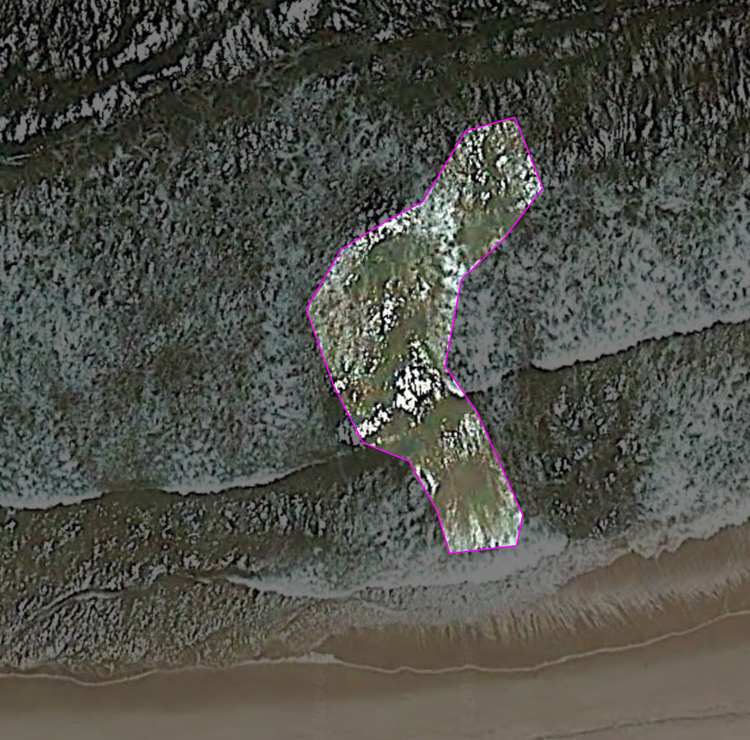} & 
         \includegraphics[width=0.19\textwidth]{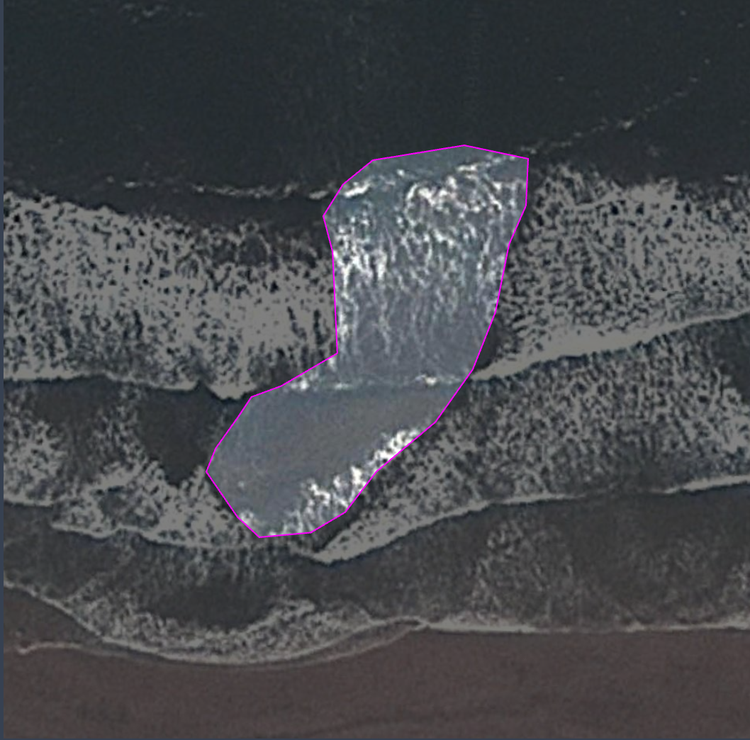} &  \tabularnewline
         \begin{turn}{90} {\raggedright Without rips} \end{turn}
         \includegraphics[width=0.19\textwidth]{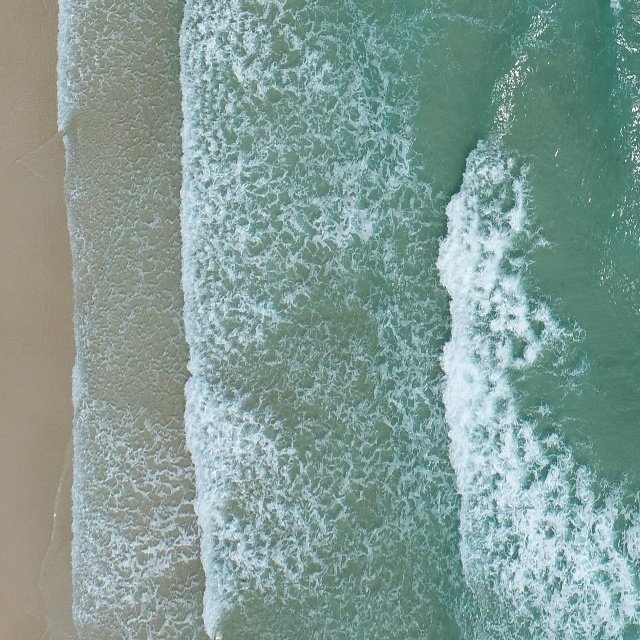} & 
         \includegraphics[width=0.19\textwidth]{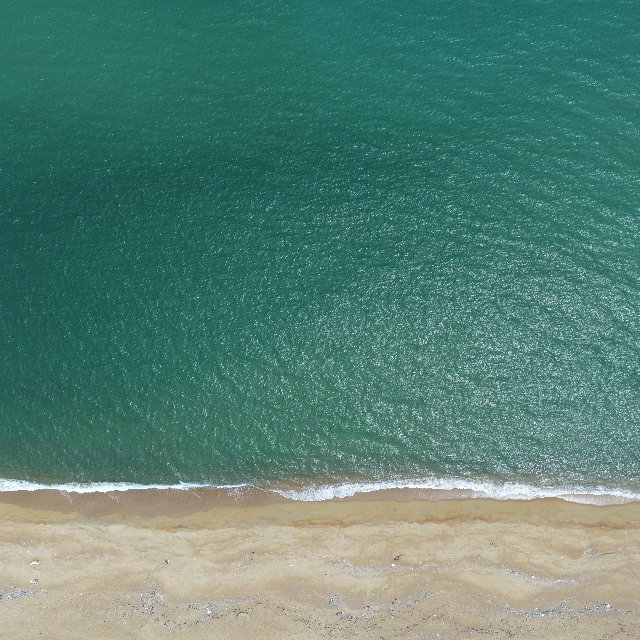} & 
         \includegraphics[width=0.19\textwidth]{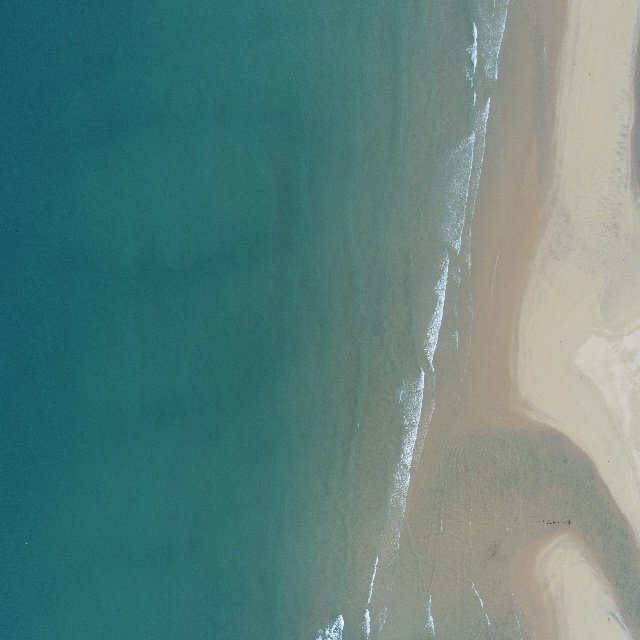} & 
         \includegraphics[width=0.19\textwidth]{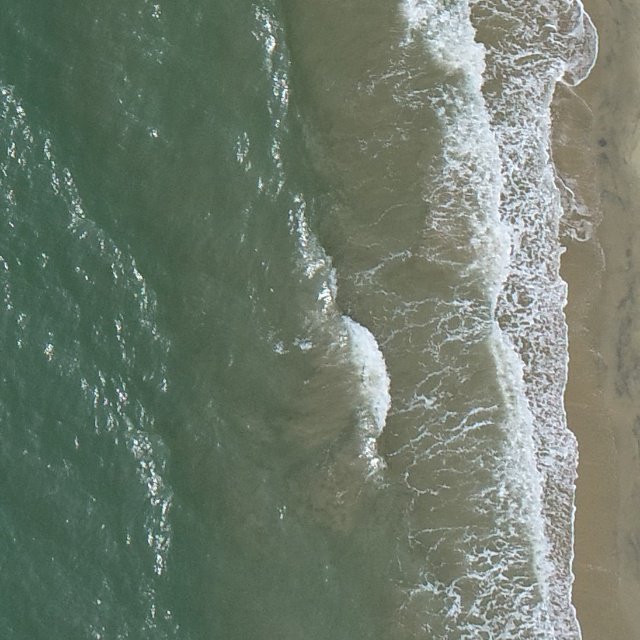} & 
         \includegraphics[width=0.19\textwidth]{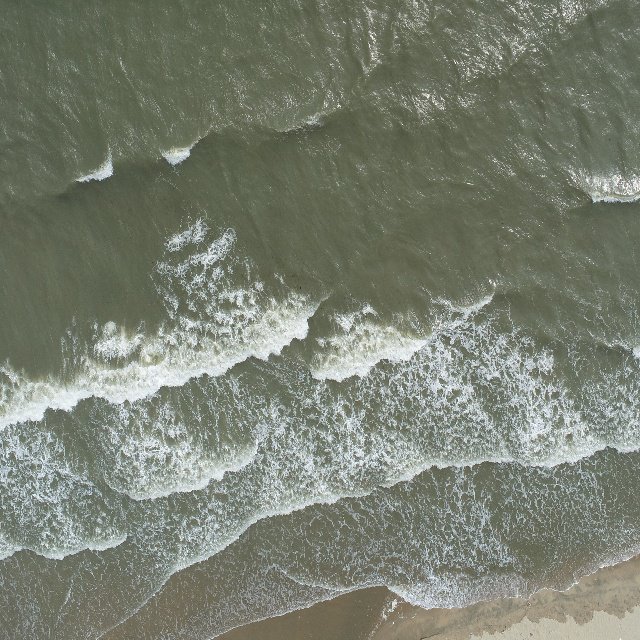} &  
    \end{tabular}
    \captionsetup{justification=justified}
    \caption[Images from the training dataset]{Selected images from the training dataset with corresponding ground-truth annotations. Rip currents exhibit an amorphous nature, resulting in varying appearances even for the same rip current, depending on the snapshot's timing and surrounding conditions. The top and middle rows display rip currents that are easy and difficult to spot, respectively. The bottom row showcases images without rip currents in various settings.}
    \label{fig:images_train}
\end{figure*}
Rip current detection has been analyzed by a Rashid~\etal\cite{rashid2020ripnet} in their proposed RipNet architecture. The authors approach  rip current detection as an anomaly detection problem rather than an object detection or classification problem, improving their performance through an autoencoder-based anomaly detection, reducing the need of additional negative samples during training, and achieving a considerable increase in accuracy, specificity and sensitivity metrics on the same dataset as Maryan~\etal\cite{maryan2019machine}.

The same authors proposed RipDet\cite{rashid2021ripdet}, a rip current detector based on Tiny-YOLOv3, pre-trained on the COCO-2017 dataset \cite{cocodataset} and fine-tuned on the same rip current dataset, expanded by augmentations. They obtain a significant 98.13\% detection rate on the proposed benchmark, with reduced model dimensions and increased inference speed. They continued with a proposal of RipDet+\cite{rashid2023reducing}, which improves the previous architecture by adding residual blocks. Similar to RipDet, RipDet+ uses the weights of Tiny-YOLOv3 pre-trained on the COCO-2017 dataset for initializing the weights of all layers, except in residual and penultimate layers. Afterwards, the entire model is fine-tuned on the rip current dataset of 4,192 images. This strategy increased their detection accuracy to 98.55\%.

De Silva~\etal\cite{desilva2021frcnn} introduced a new dataset, mainly collected from Google Earth, that includes images of rip currents and similar beaches without rip currents. They used this dataset to train a Faster R-CNN\cite{ren2015faster} model, which was tested on video data. Their approach also involves a custom temporal aggregation stage that leverages the continuity of video data to improve detection accuracy and reduce bounding box variance. The authors demonstrated the superiority of their method compared to previous approaches. Although the exact results are credible, they are not entirely reproducible, since the ground-truth annotations of the test dataset is not known. Moreover, the validation results are not provided.

Zhu~\etal\cite{zhu2022yolo} expanded the training dataset collected by de Silva~\etal\cite{desilva2021frcnn} with additional images from the internet and South China coastal beach sites. They developed a YOLO-Rip network based on the YOLOv5s architecture, adding modifications to improve rip current detection. Their network showed increased accuracy on their dataset compared to YOLOv5, as well as faster training speeds than previous implementations.

Our method advances existing approaches by employing instance segmentation for more precise rip current detection, utilizing a comprehensive dataset to enhance generalization, and leveraging the YOLOv8 architecture for faster and more accurate results. These improvements enable detailed analysis and facilitate real-time applications, ultimately contributing to enhanced beach safety measures.

\subsection{Rip Current Detection Datasets}
The development of rip current detection methods, particularly those based on machine learning and computer vision, relies heavily on the availability of high-quality, annotated datasets. In this section, we review some of the notable datasets that have been assembled and utilized for rip current detection research. 

Maryan~\etal dataset\cite{maryan2019machine}: this dataset contains 514 rip channel examples, including the test dataset. The rip channels are images of $24 \times 24$ pixels  extracted from images of $1334 \times 1334$ pixels. These larger images are timex images downloaded from the backlog of beach imagery on the Oregon State University website\cite{osuwebsite}. The images have been orthorectified and are time-averaged over 1,200 frames collected at 2 Hz over 10 minutes. The rip channel samples are extracted from these larger images using the GIMP image editor and normalized to $24\times24$ pixels. The samples are also converted to grayscale to minimize the effects of different lighting conditions on the detection models. To create a larger dataset suitable for training convolutional neural networks, data augmentation is applied, resulting in a dataset of 4,000 rip channel images. The dataset is used for training and evaluating various rip current detection algorithms in several studies, such as those conducted by Rashid~\etal\cite{rashid2020ripnet, rashid2021ripdet, rashid2023reducing}.

De Silva~\etal dataset\cite{desilva2021frcnn}: the primary source for the dataset was Google Earth, which provided high-resolution aerial images of rip currents and non-rip current beach scenes. In total, the dataset consists of 1,740 rip current images and 700 non-rip current images, with sizes ranging from $1086 \times 916$ to $234 \times 234$ pixels. Ground-truth annotations with axis-aligned bounding boxes were added to the rip current images. This dataset was used for training the models described in the paper. In addition to the static image dataset, De Silva~\etal collected a test dataset consisting of 23 video clips, totaling 18,042 frames. Among these, 9,053 frames contain rip currents, and 8,989 frames do not. The image size varies from $1280 \times 720$ to $1080 \times 920$ pixels. A co-author, who is also a rip current expert at NOAA, verified the ground-truth annotations for this dataset. The frames of this video dataset were used for testing the models. It is important to note that the static images in the training set were taken from high elevation, while the test videos were taken from a lower perspective. Bounding box annotations are provided for the 1,740 rip current images but not for the videos.

YOLO-Rip dataset\cite{zhu2022yolo}: the YOLO-Rip authors expanded the dataset provided by de Silva~\etal by collecting several sets of real beach scene photographs along the coast of South China. They selected 1,352 high-resolution images, 746 containing rip currents and 606 without, ranging in size from $4000 \times 2250$ to $480 \times 360$ pixels. Axis-aligned bounding boxes were used to label the rip boundaries in the images containing rip currents. This expanded dataset aimed to improve the model's ability to accurately identify rip currents from diverse types of images and enhance its real-world applicability.

\section{Proposed Benchmark Dataset}
Rip currents exhibit amorphous characteristics, lacking a definite shape. Existing datasets predominantly consist of bounding box annotations, which, despite being more convenient to create, provide only a rough estimation of the rip current channel location. This often results in either the inclusion of noise or the exclusion of relevant information. Our approach builds upon the datasets provided by Zhu~\etal and De Silva~\etal. Similar to De Silva~\etal, we divide the dataset into training images and test videos, supplying files and annotations for both.
\begin{table*}[t]
  \centering
    \begin{tabular}{@{}lcccccccccc@{}}
        \toprule
        \multirow{2}{*}{Model} & Last  & Best  & Train  & \multirow{2}{*}{FPS} & \multirow{2}{*}{Precision} & \multirow{2}{*}{Recall} & \multirow{2}{*}{mAP50} & \multirow{2}{*}{mAP50:95} & Stddev \\
         & epoch & epoch  & time (h) &  &  &  &  &  & (mAP50) \\
        \midrule
        YOLOv8n-seg & 180.8 & 130.8 & 1.58 & 73.33  & 88.27 & 84.67 & 88.94 & 45.75 & 2.09 \\
        YOLOv8s-seg & 167.3 & 117.3 & 1.83 & 33.74 & 87.62 & 84.67 & 89.25 & 46.66 & 1.95\\
        YOLOv8m-seg & 147.2 & 97.20 & 2.23 & 15.54 & 88.04 & 85.04 & 89.92 & 47.48 & 2.23\\
        YOLOv8l-seg & 135.6 & 85.6 & 2.91 & 9.664 & 89.04 & 84.97 & 89.72 & 46.85 & 1.90\\
        YOLOv8x-seg & 139.6 & 89.6 & 4.86 & 6.324  & 88.74 & 84.26 & 90.19 & 47.45 & 1.45\\
        \bottomrule
    \end{tabular}
  \caption{Validation results on the testing dataset. Each metric value is the average over all 10 folds, except for stddev. The difference in mAP50 (our most relevant metric) is marginal between the models, even when comparing the smallest to the largest. However, as the model size increases, the FPS rate drops drastically. } 
  \label{tab:validation_results}
\end{table*}
\subsection{Training images}
We have annotated 2,466 images from the YOLO-Rip dataset for instance segmentation. As rip currents are amorphous, we assumed an approximate shape based on expert knowledge in cases where the entire shape was not visibly apparent. Additionally, we included 1,307 images of similar beaches without rip currents from the same dataset.

\textbf{Scenes:} numerous images in the dataset are satellite views of beaches, captured from Google Earth. While most images display a clear wave-breaking pattern, rip currents occasionally reveal their shape through distinct water colors. Rip currents, defined by their outward-flowing sea channels, exhibit periodic apparent shapes due to continuous water movement and waves. In videos, the rip current channel's shape is more easily discernible, thanks to extended observation. However, in images, the rip current channel's visibility depends on the snapshot's timing, which can vary from obvious to obstructed. In the latter case, informed assumptions are made based on the shape.

\textbf{Annotations:} annotations were generated using Roboflow's polygon tool\cite{roboflow2022}. One of the challenges in rip current segmentation is the annotation stage. As previously mentioned, images showcase rip currents at different moments and with varying apparent shapes. Lacking context, assumptions must be made. Depending on the image and the rip current's apparent shape, our approach varied between annotating the obvious shape, the assumed channel, or a combination of both to provide as much information as possible to the model. Some images contain a significant amount of calm offshore water, devoid of easily distinguishable characteristics. In these instances, decisions were made regarding where to terminate the rip current segmentation based on the image itself. Occasionally, the rip current's outer boundaries are segmented by waves before breaking, while at other times, waves are included in the segmentation due to the rip current's apparent shape in front of and behind the waves. The annotations were checked by a co-author who is an expert in the field of rip currents.

\subsection{Test videos}
We have compiled 17 videos over three years (2021, 2022, and 2023), captured at 30 FPS, resulting in a total of 28 minutes and 58 seconds of aerial footage of rip currents and 24,295 frames. Among these videos, 2 are recorded at a resolution of $3840\times2160$, while the rest at a resolution of $1920\times1080$. We recorded the videos at various dates and times from a Black Sea beach site in Eforie Nord, Romania. 

The duration of the videos range from as short as 2 seconds to as long as 10 minutes and 46 seconds. This variation is intentional to evaluate the model's performance on both short and long instances, since it was only trained on images. The average duration of the videos is 1 minute and 42 seconds, with a median value of 1 minute. Most videos feature a hovering fixed standpoint overlooking the rip current, with only minor adjustments in position and angle due to wind. Some footage captures either the camera or the drone moving, taken from different elevations. The $1920\times1080$ videos were recorded at f2.8 and ISO 100.
We also included 5 beachfront videos without rip currents from De Silva~\etal test data in order to evaluate for false positives, as well.

We encourage other researchers and organizations to contribute to the collection of diverse rip current samples, fostering collaboration and expansion of the dataset for improved model performance.

\section{Baseline method: YOLOv8}
YOLOv8\cite{Jocher_YOLO_by_Ultralytics_2023} is the latest version in the YOLO (You Only Look Once) series of object detection algorithms, known for their real-time detection capabilities and accuracy \cite{redmon2016you, redmon2016yolo9000, redmon2018yolov3, bochkovskiy2020yolov4, Jocher_YOLOv5_by_Ultralytics_2020, li2022yolov6, li2023yolov6, wang2022yolov7}. YOLOv8 employs an anchor-free detection mechanism, an enhanced feature pyramid network, and a modified loss function to achieve improved performance over its predecessors. It is designed to segment objects in images by predicting bounding boxes and associated class probabilities using a single convolutional neural network (CNN).
Before choosing YOLOv8, we considered other models that are in the top ranking of COCO instance segmentation test-dev, such as EVA \cite{fang2022eva}, 	
FD-SwinV2-G \cite{wei2022contrastive}, BEiT-3 \cite{wang2022image}, MasK Dino \cite{li2022mask}, ViT-Adapter-L \cite{chen2022vision} and SwinV2-G \cite{liu2022swin}, as well as older methods, such as RDSnet \cite{wang2020rdsnet}, TensorMask \cite{chen2019tensormask}, PolarMask \cite{xie2020polarmask}, SipMask \cite{cao2020sipmask}, D2Det \cite{cao2020d2det} and Mask R-CNN \cite{he2017mask}. However, we selected YOLOv8 as our baseline method for several reasons. First, YOLO algorithms have consistently demonstrated strong performance in object detection tasks, making them competitive and suitable for our rip current instance segmentation problem. The real-time detection capabilities of YOLOv8, in particular, are essential for applications involving public safety and emergency response, as demonstrated by the success of previous YOLO versions in various object detection tasks. Second, YOLOv8 is a well-established method with an extensive user community, providing accessible implementation resources and making it an appropriate choice for a benchmark study. Third, it is easy to deploy and use in multiple version sizes.

In our study, we opted to use YOLOv8 out-of-the-box without any modifications. This decision ensures a clear and unbiased evaluation of our dataset, and provides a straightforward starting point for future research. By using YOLOv8 unmodified, we establish a performance baseline that can be easily replicated and compared with other methods applied to our dataset.

To determine the most suitable model size for our experiments, we conducted a comparative analysis of the various YOLOv8 models. This comparison helps to identify the trade-offs between detection accuracy and computational efficiency, ensuring that the selected model is appropriate for the specific hardware and application scenario. We present a detailed analysis of the model sizes and their performance on our dataset in the experiments section.

\section{Experiments}
\subsection{Environment and parameters}
The training and inference were run on a 24GB NVIDIA GeForce RTX 3090 GPU, using YOLOv8.0.53, Python 3.10.4, PyTorch 1.12.1 and CUDA 11.7. We trained on images of $640 \times 640$ pixels for 300 epochs, on mini-batches ranging from 64 to 12, depending on model size, with an early stopping patience parameter of 50 epochs. The models stopped before overfitting somewhere between epoch 104 and 214. The detailed model parameters and logs can be accessed in the github repository.

\subsection{Evaluation metrics}
For evaluation on the validation data, we take a look at several metrics: last epoch, to see where the model training stopped due to overfitting (with a patience of 50 epochs), best epoch, \ie the epoch with the best results before starting to overfit, training time (average per epoch), frames per second (FPS), precision \eqref{eq:precision}, recall \eqref{eq:recall}, mAP50, map50:95 \eqref{eq:mAP}, and standard deviation over the 10 folds for the mAP50 values. All values, except for the standard deviation, are the average over all 10 folds.
We use the mean Average Precision (mAP) as our primary evaluation metric, specifically mAP50. We also look at frames per second, considering real-world applications. The mAP is based on the Intersection over Union (IoU) \eqref{eq:iou}, which is derived from the Jaccard index, and represents the average of the precision values at different recall levels for a given IoU threshold.

The IoU, or Jaccard index, measures the similarity between two sets and is defined as the ratio of their intersection to their union:
\begin{equation}\label{eq:iou}
IoU(A, B) = \frac{|A \cap B|}{|A \cup B|}.
\end{equation}



\begin{table*}[t]
  \centering
    \begin{tabular}{@{}ccccccc@{}}
        \toprule
        Video & Frames & YOLOv8-n (\%)& YOLOv8-s (\%)& YOLOv8-m (\%)& YOLOv8-l (\%)& YOLOv8-x (\%) \\
        \midrule
        DJI\_0092.mp4 & 278 & 100.0 & 75.90 & \textcolor{red}{10.43} & 100.0 & \textcolor{red}{0.720} \\
        DJI\_0093.mp4 & 646 & 96.90 & 72.45 & 97.52 & 100.0 & 96.75 \\
        DJI\_0094.mp4 & 5248 & 50.84 & 71.55 & 91.54 & 89.81 & 84.81 \\
        DJI\_0095.mp4 & 2516 & 61.29 & \textcolor{orange}{30.45} & 76.27 & 56.68 & \textcolor{red}{18.76} \\
        DJI\_0384.mp4 & 4304 & 93.66 & 76.65 & 66.94 & 72.10 & 85.04 \\
        DJI\_0393.mp4 & 84 & 100.0 & 100.0 & 100.0 & 100.0 & 100.0 \\
        DJI\_0394.mp4 & 3623 & 71.93 & 86.28 & 93.43 & 85.04 & 97.90 \\
        DJI\_0395.mp4 & 4583 & 75.98 & 98.19 & 95.99 & 87.67 & 92.84 \\
        DJI\_0410.mp4 & 1397 & 94.56 & \textcolor{orange}{47.24} & 95.49 & 59.06 & 85.25 \\
        DJI\_0423.mp4 & 1825 & 65.04 & 81.04 & \textcolor{orange}{47.18} & \textcolor{red}{14.85} & \textcolor{orange}{46.79} \\
        DJI\_0438.mp4 & 1238 & 81.83 & 70.19 & 92.08 & 52.18 & 57.27 \\
        DJI\_0448.mp4 & 10 & 100.0 & \textcolor{red}{10.00} & \textcolor{red}{10.00} & 100.0 & 100.0 \\
        DJI\_0670.mp4 & 1812 & 76.49 & \textcolor{orange}{46.52} & \textcolor{orange}{46.30} & 56.73 & 74.78 \\
        DJI\_0678.mp4 & 1238 & 89.74 & 98.87 & 91.68 & \textcolor{orange}{22.29} & 99.35 \\
        DJI\_0679.mp4 & 634 & 98.58 & 100.0 & 100.0 & 97.16 & 99.37 \\
        DJI\_0680.mp4 & 1359 & 83.08 & 93.97 & 100.0 & 91.39 & 89.04 \\
        DJI\_0808.mp4 & 2194 & 54.51 & \textcolor{red}{0.270} & \textcolor{red}{0.000} & \textcolor{red}{8.200} & \textcolor{red}{0.730} \\
        no\_rip\_01.mp4 & 939 & 98.94 & 89.03 & 97.23 & 93.29 & 95.63 \\
        no\_rip\_02.mp4 & 924 & 100.0 & 100.0 & 100.0 & 100.0 & 100.0 \\
        no\_rip\_03.mp4 & 921 & \textcolor{red}{12.49} & \textcolor{orange}{27.25} & 92.40 & \textcolor{orange}{31.38} & \textcolor{red}{18.13} \\
        no\_rip\_04.mp4 & 926 & 82.07 & 50.76 & 89.09 & 90.71 & 85.85 \\
        no\_rip\_05.mp4 & 922 & 82.21 & 80.91 & 73.21 & 88.61 & 81.24 \\
        no\_rip\_06.mp4 & 451 & 93.57 & \textcolor{orange}{49.22} & 72.73 & 73.61 & \textcolor{orange}{49.67} \\
        no\_rip\_07.mp4 & 900 & 83.67 & 93.44 & 89.00 & \textcolor{orange}{48.67} & 54.00 \\
        no\_rip\_11.mp4 & 2576 & 82.92 & 52.60 & 78.22 & 76.36 & 74.18 \\
        \midrule
        All videos & (macro avg) & \textcolor{blue}{81.21} & 68.11 & 76.27 & 71.83 & 71.52 \\
        All videos & (micro avg) & 74.83& 69.43 & \textcolor{blue}{78.86} & 69.62 & 73.30 \\
        \bottomrule

    \end{tabular}
  \caption{Results on the test dataset (videos) for rip current segmentation. Values represent the percentage of successfully segmented frames using mAP50. Macro and micro averages are shown in the last two rows. Best results are highlighted in blue, extreme failure cases (	$<$20\% accuracy) in red, and 20\%-50\% accuracy in orange.}
  \label{tab:test_results}
\end{table*}

Precision and Recall are calculated as follows:
\begin{equation}\label{eq:precision}
Precision = \frac{True\ Positives}{True\ Positives + False\ Positives},
\end{equation}
\begin{equation}\label{eq:recall}
Recall = \frac{True\ Positives}{True\ Positives + False\ Negatives}.
\end{equation}

In general, the Average Precision (AP) for a single class is calculated by ranking the model's predictions by their confidence scores, and then computing the area under the precision-recall curve:
\begin{equation}\label{eq:AP}
AP = \sum_{n} (Recall_n - Recall_{n-1}) \cdot Precision_n.
\end{equation}

Since our model detects only one class (rip currents), the mAP is equal to the AP of that class. For mAP50, we compute the AP at an IoU threshold of 0.5. For mAP50:95, we calculate the AP for each IoU threshold from 0.5 to 0.95 in increments of 0.05, and then average those values:
\begin{equation}\label{eq:mAP}
mAP\!=\!\frac{AP_{IoU=0.5} + AP_{IoU=0.55} + \cdots + AP_{IoU=0.95}}{k},
\end{equation}
where $k$ is the number of IoU thresholds considered.

By using mAP50 and mAP50:95, we evaluate our model's ability to accurately segment rip currents at varying IoU thresholds. We also assess the model's efficiency by considering training speed and inference time, ensuring that our solution is practical for real-world applications in rip current detection and analysis.

For evaluation on the test videos, we examine the number of frames with accurately segmented rip currents using the mAP50 metric. We assess the accuracy for each video and analyze the failure cases. The final results are presented as both macro average, considering the average accuracy across all videos, and micro average, taking into account the number of true positives across all frames.

\begin{figure*}
\centering
\setlength{\tabcolsep}{1pt}
\begin{tabular}{c c c c }
     (a) & (b) & (c) & (d) \tabularnewline
     \begin{turn}{90} {\raggedright True Positives} \end{turn}
     \includegraphics[width=0.24\textwidth]{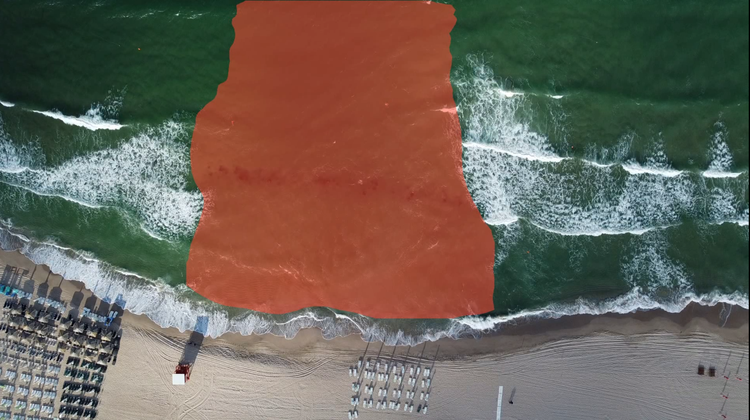} & 
     \includegraphics[width=0.24\textwidth]{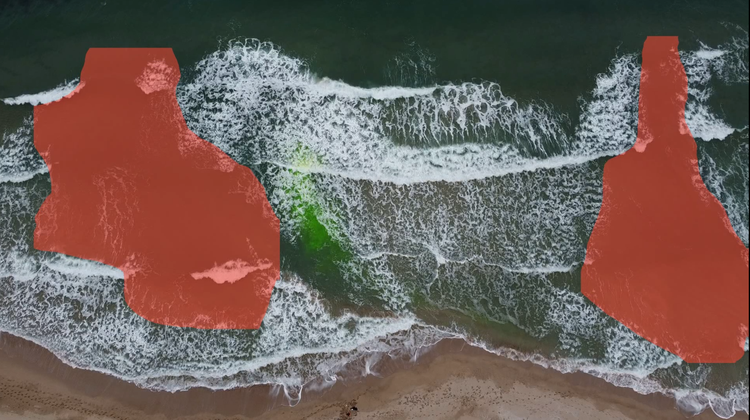} & 
     \includegraphics[width=0.24\textwidth]{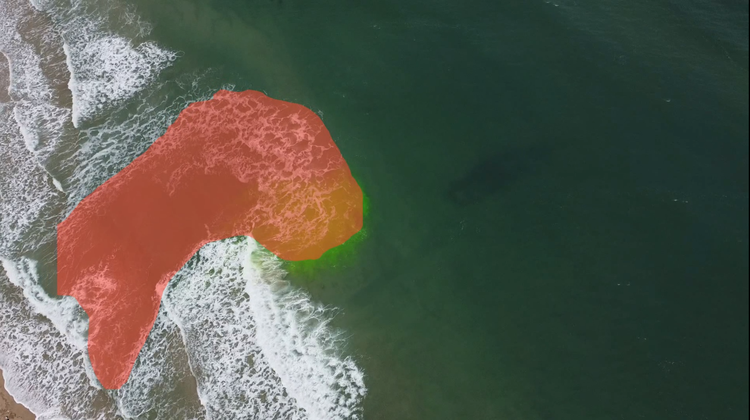} & 
     \includegraphics[width=0.24\textwidth]{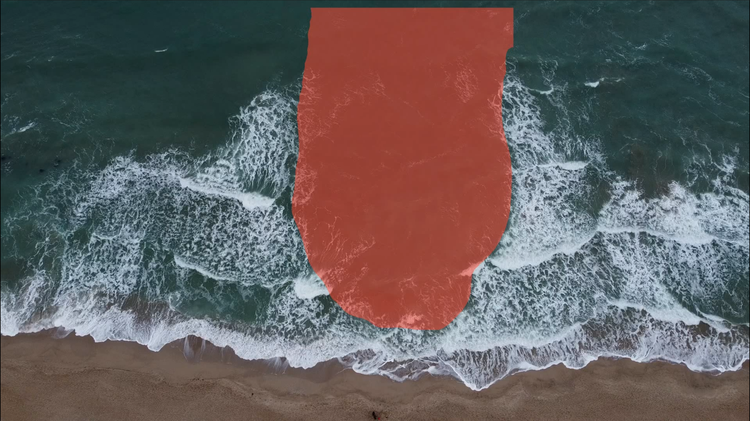} 
     \tabularnewline
     \begin{turn}{90} {\raggedright True Negatives} \end{turn}
     \includegraphics[width=0.24\textwidth]{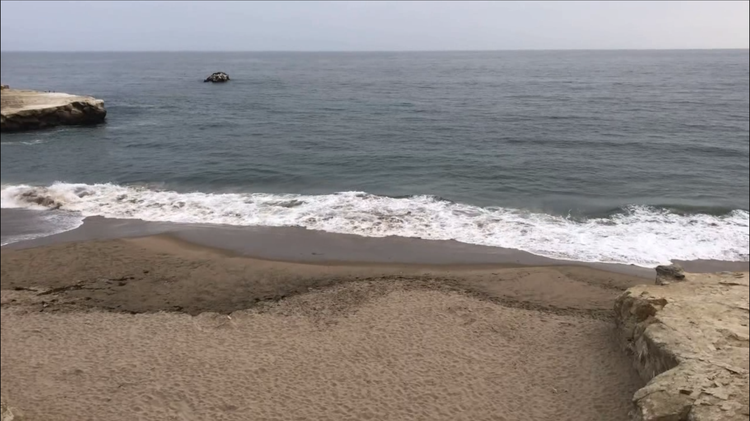} & 
     \includegraphics[width=0.24\textwidth]{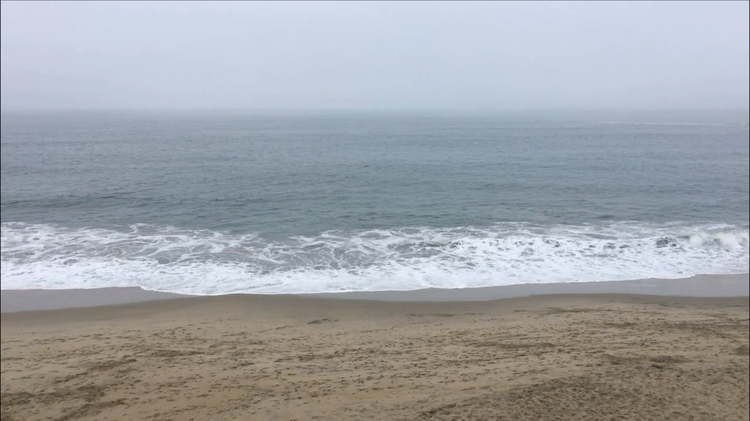} & 
     \includegraphics[width=0.24\textwidth]{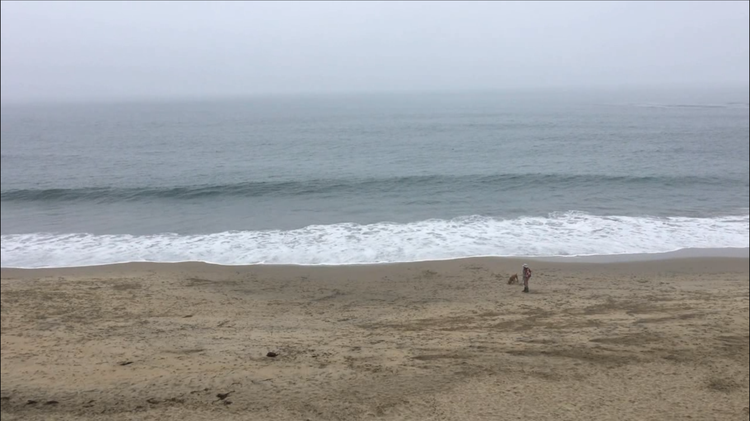} & 
     \includegraphics[width=0.24\textwidth]{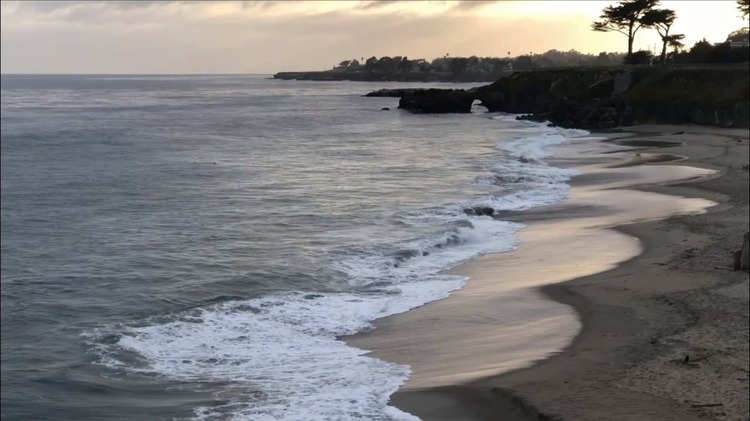} 
     \tabularnewline
     \begin{turn}{90} {\raggedright False Positives} \end{turn}
     \includegraphics[width=0.24\textwidth]{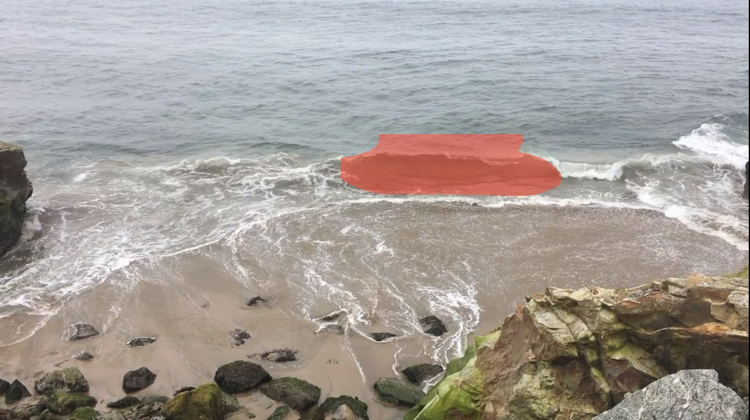} & 
     \includegraphics[width=0.24\textwidth]{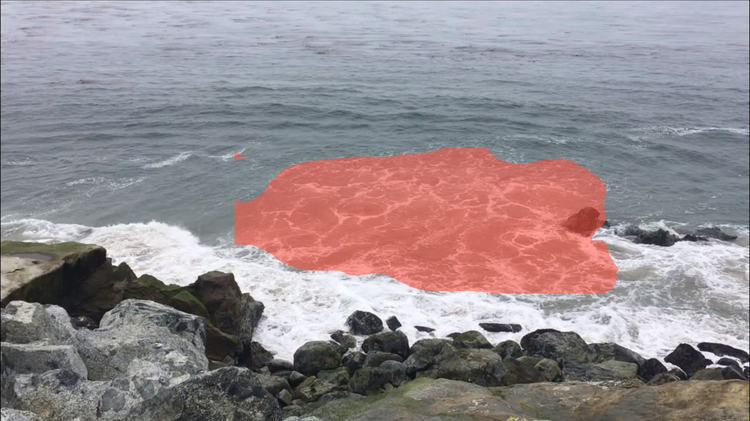} & 
     \includegraphics[width=0.24\textwidth]{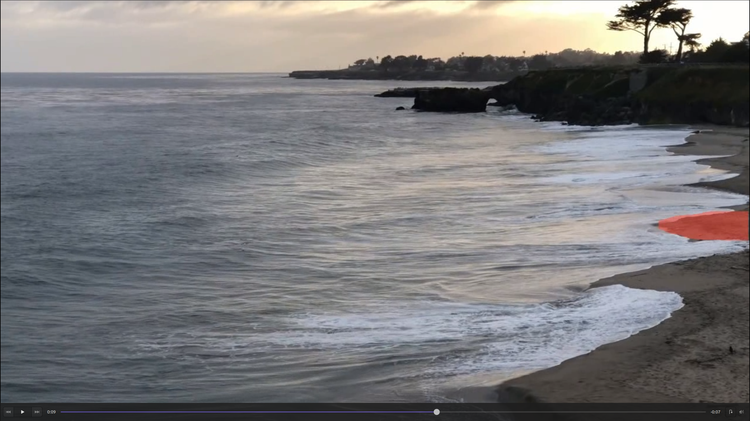} & 
     \includegraphics[width=0.24\textwidth]{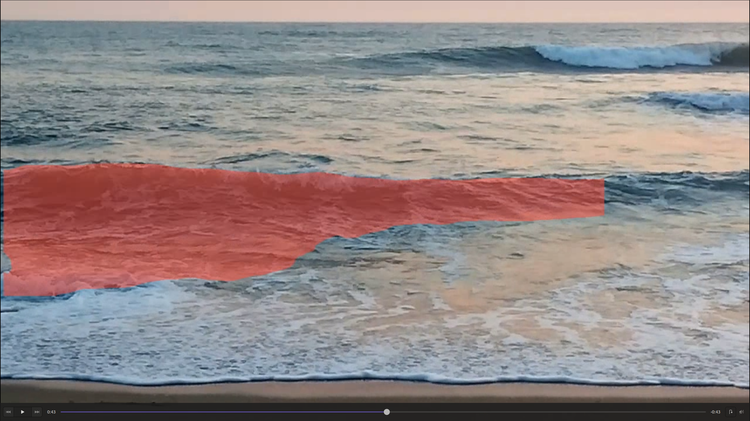} 
     \tabularnewline
     \begin{turn}{90} {\raggedright False Negatives} \end{turn}
     \includegraphics[width=0.24\textwidth]{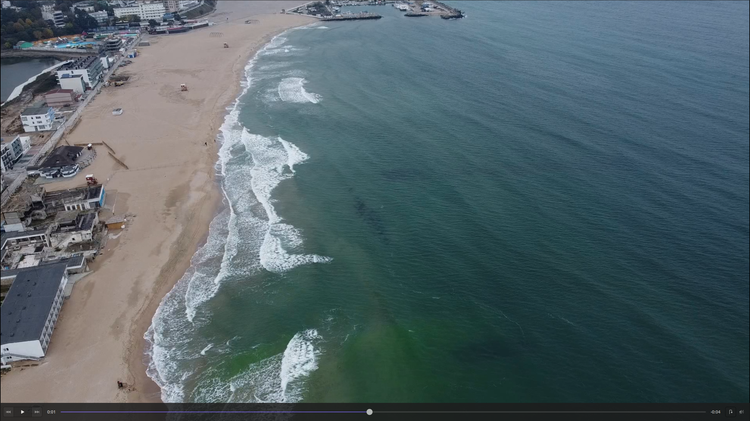} & 
     \includegraphics[width=0.24\textwidth]{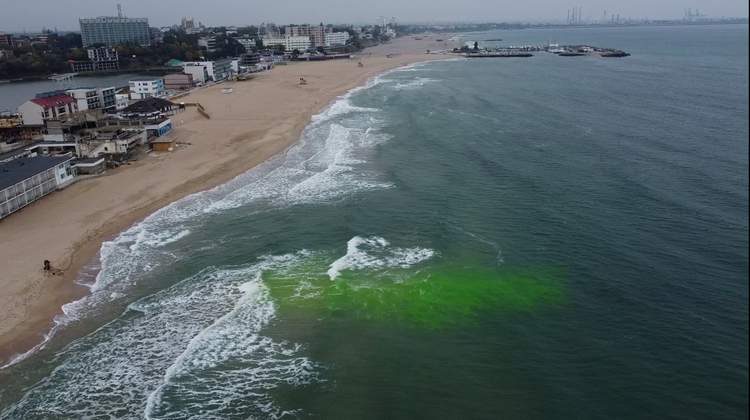} & 
     \includegraphics[width=0.24\textwidth]{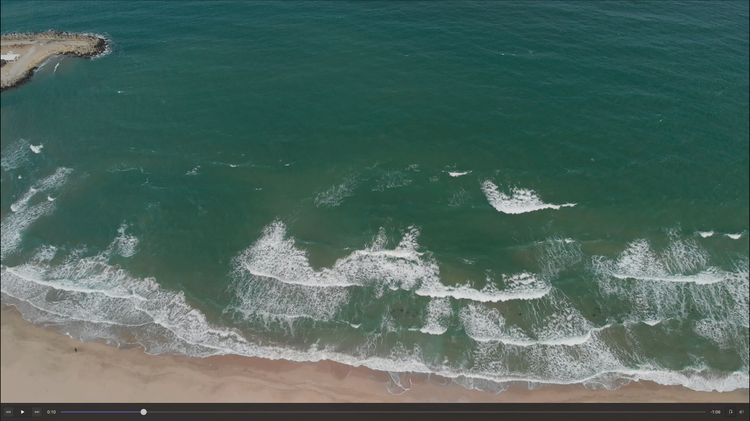} & 
     \includegraphics[width=0.24\textwidth]{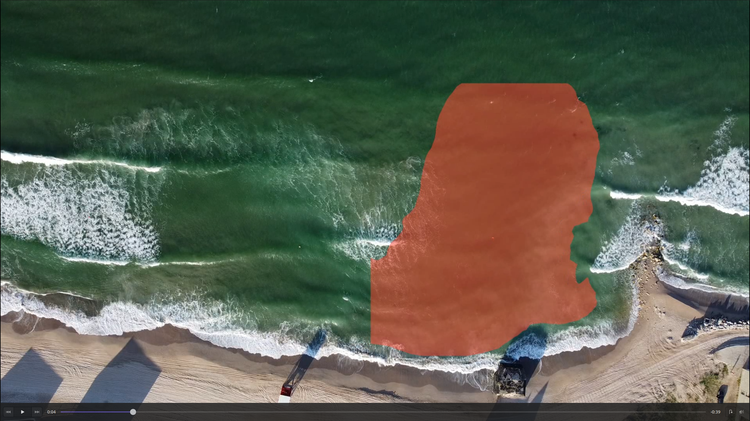} 
     \tabularnewline
\end{tabular}
  \caption{Example of results on testing dataset (using the nano model). On the first two rows we have the correct predictions, the true positives and the true negatives. On the last row we have the incorrect predictions, the false positives and the false negatives. Notice column (d) on the last row, where the model manages to correctly predict one rip current, but misses the second one.}
  \label{fig:testing_results}
\end{figure*}
\subsection{Training Method}
Given the diverse nature of our dataset, which includes satellite, aerial, and beachfront images, we employed a k-fold cross-validation technique with 10 folds and a 90\% train and 10\% validation split. The dataset comprises 2,466 images of rip currents and 1,307 images without rip currents. To avoid an imbalanced fold, we separately applied a 10-fold cross-validation on each category (rip and non-rip), and then concatenated the file lists before training. This approach, known as stratified sampling, ensures a consistent class distribution in each fold.

For full reproducibility, the code and files used in each fold are provided in the github repository. We trained all model sizes (n, s, m, l, and x) on the same 10-fold split, comparing all the results, training and inference times, but focusing on mAP50 and FPS. The metrics compared for each model represent the average over all 10 folds, except for standard deviation, which is the metric for the standard deviation of the mAP50 values between all 10 folds.

\subsection{Results}
We compared all the models via cross-validation on images and testing on videos. 
The validation results and speeds can be seen in Table \ref{tab:validation_results}. On the training and validation data, as expected, the difference in FPS or inference time between models is significant and correlates with size. Contrary to initial expectations, the larger models do not yield a drastic improvement in segmentation performance on the validation data.

The test results can be seen in Table \ref{tab:test_results}, and manually selected frames have been included in Figure \ref{fig:testing_results}. While the results vary, we report a maximum macro average accuracy of 81.21\% on all videos and a maximum micro average accuracy of 78.86\% on all frames. These results could be improved with several post-processing methods, such as taking into consideration the temporal dimension of the videos. It is interesting that the best performing model in most situations is the smallest one. Zeiler and Fergus\cite{zeiler2014visualizing} have shown that deeper layers of CNNs tend to learn more and more complex patterns. Seeing that we do not detect an exact shape, but more of a general pattern (due to their amorphous property), it is possible that this problem is easier to solve with low-level features.

The nano model successfully detects the rip currents in many situations (more than 50\% of the frames of each video), leading to a clear segmentation with minimal inclusion of noise and minimal exclusion of relevant information, even in situations where multiple rip currents are visible. For the tested videos, the false negatives are less than 20\% of the frames, depending on the video. To all these, there is only one clear exception: the video no\_rip\_03.mp4. This video is taken at around head-level elevation. The false positive is detected in the horizontal breaking pattern of the waves. This is an important error, as the high number of false positives makes it harder to correct even by post-processing the information. An in-depth analysis of false negatives reveals that certain factors, such as wave patterns and lighting conditions, could be responsible for these errors. Future work should focus on addressing these factors by fine-tuning model parameters and incorporating additional context information.

\section{Conclusion}

In this study, we introduced a novel training dataset for rip current instance segmentation, a task not previously approached using deep learning. We also presented 17 videos with both bounding box and instance segmentation annotations, paving the way for methods leveraging the temporal continuity of videos. This dataset is the first of its kind. We employed the YOLOv8 model in all available sizes (n, s, m, l, and x) and compared their training speed, inference speed, mAP50, and mAP50:95 performance. The best performing model is the smallest one (nano), likely due to the amorphous nature of rip currents. The nano model performs well on all videos except one, where it yields a strong false negative. Addressing false negatives is crucial for user safety. This study demonstrates the potential of deep learning models, specifically YOLOv8, for rip current segmentation. However, we acknowledge the limitations of our approach due to the dataset distribution, and the fact that the model can only generalize based on the available data. 

In future work, we aim to focus on expanding the dataset, investigating alternative architectures, fine-tuning model parameters, developing real-time detection systems for videos, including temporal information, and adding multiple classes such as swimmers, surfers, and boats in the segmentation task. The developed model holds promise for real-world applications, such as integration into beach safety systems or mobile devices for user alerts. 
Further work is required to optimize the model for deployment in these scenarios. Moreover, we plan to explore collaboration opportunities with domain experts, organizations, and beach authorities to improve data collection and model deployment in real-world environments.

The presented dataset, models, and evaluation methods aim to advance the field and improve beach safety by preventing rip current-related incidents. To facilitate future research, we have made the code, logs, exact k-fold splits, and datasets publicly available.

\section*{Acknowledgments}
This work was partly supported by the Humboldt Foundation.

We thank previous authors for their research and dataset contributions. Through continued effort, a larger, diversified dataset is being developed. Rip current detection has life-saving potential, and each study contributes to enhancing global beach safety.

{\small
\bibliographystyle{ieee_fullname}
\bibliography{egbib}
}

\end{document}